\documentclass[11pt,a4paper]{article}
\usepackage[hyperref]{acl2020}
\usepackage{times}
\usepackage{latexsym}
\usepackage{graphicx}
\usepackage{amsmath}
\usepackage{amsfonts}
\usepackage{amssymb}
\usepackage{subfig}

\usepackage{microtype}
\aclfinalcopy % Uncomment this line for the final submission
\usepackage{float} %for image formatting

\title{Towards More Accurate Prediction of Human Empathy and Emotion in Text and Multi-turn Conversations by Combining Advanced NLP, Transformers-based Networks, and Linguistic Methodologies}
\author{\\Manisha Singh (\texttt{manishas@uw.edu}), Divy Sharma (\texttt{divy@uw.edu}), \\Alonso Ma (\texttt{amatake@uw.edu}), Nora Goldfine  (\texttt{ngoldf@uw.edu})}
\begin{document}

\maketitle

\begin{abstract}
Based on the WASSA 2022 Shared Task on Empathy Detection and Emotion Classification, we predict the level of empathic concern and personal distress displayed in essays. For the first stage of this project we implemented a Feed-Forward Neural Network using sentence-level embeddings as features. We experimented with four different embedding models for generating the inputs to the neural network. The subsequent stage builds upon the previous work and we have implemented three types of revisions. The first revision focuses on the enhancements to the model architecture and the training approach. The second revision focuses on handling class imbalance using stratified data sampling. The third revision focuses on leveraging lexical resources, where we apply four different resources to enrich the features associated with the dataset. During the final stage of this project, we have created the final end-to-end system for the primary task using an ensemble of models to revise primary task performance. Additionally, as part of the final stage, these approaches have been adapted to the WASSA 2023 Shared Task on Empathy Emotion and Personality Detection in Interactions, in which the empathic concern, emotion polarity, and emotion intensity in dyadic text conversations are predicted.
\end{abstract}

% architecture overview figure (early, so it will appear at the top of the page of the System Overview section)
\begin{figure*}
\centering
\includegraphics[width=0.9\textwidth]{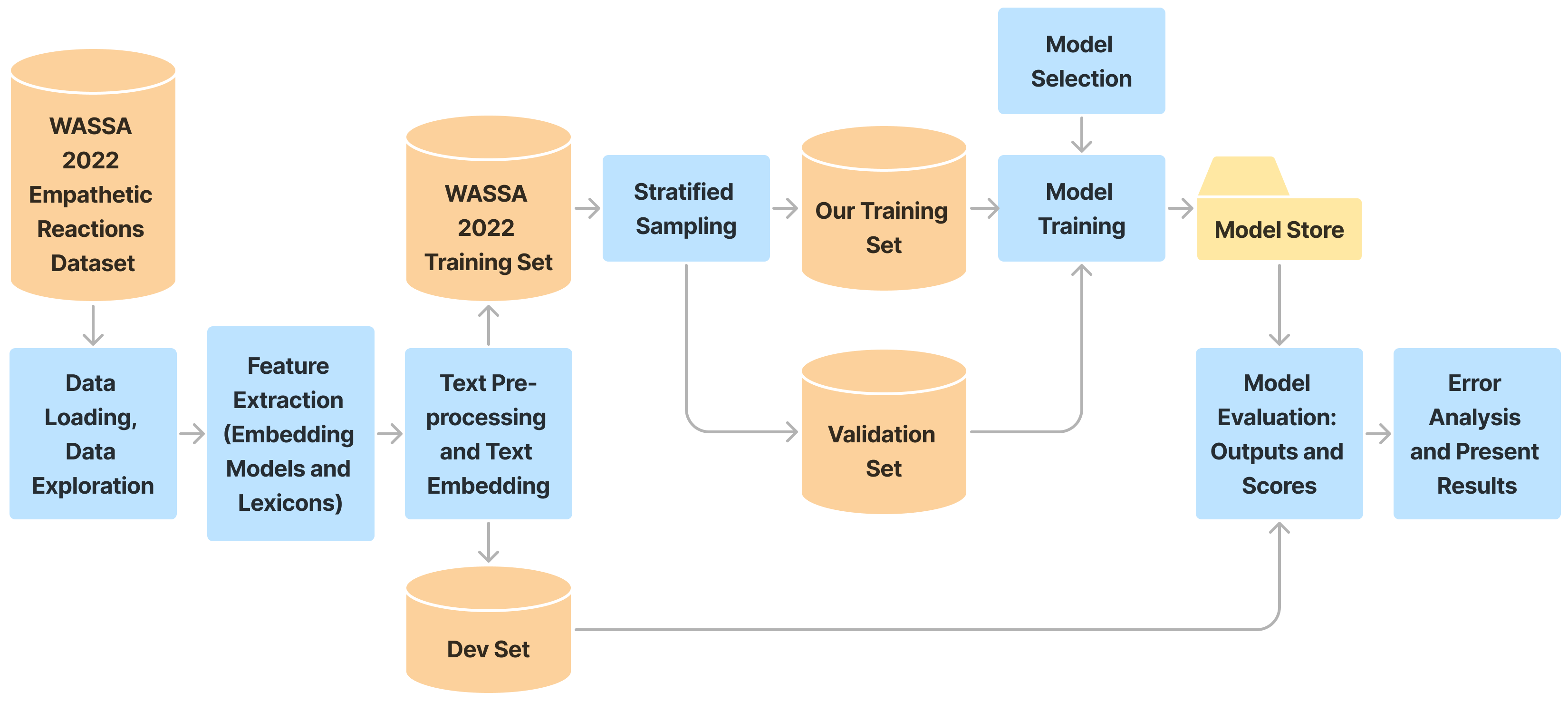}
\caption{Architecture Overview.}
\label{architecture-overview}
\end{figure*}

\section{Introduction}
As human-computer interactions increasingly integrate into our daily lives through applications, such as conversational agents where form is as critical as substance, it becomes paramount for computer systems to demonstrate natural interactions by recognizing and expressing affect. The field of Affective Computing, as proposed by \citet{picard2000affective}, aims to endow computer systems with the capability to mimic our understanding of how emotions influence human perception and behavior. This is particularly relevant in light of the fact that a vast majority of U.S. adults (86\%) receive news through digital devices such as smartphones, computers, or tablets \citep{pew-research}.
This project focuses on predicting empathy and distress elicited from news stories.

\section{Task Description}
This project is organized to address a primary task and an adaptation task. The description of the primary task is provided in Section \ref{section_primary_task_description} and the description of the adaptation task is provided in Section \ref{section_secondary_task_description}.

\subsection{Primary Task} \label{section_primary_task_description}
The primary task in this project is based on the shared task from WASSA 2022 Shared Task on Empathy Detection and Emotion Classification
\citep{buechel2018modeling}, organized at \citet{wassa-2022-codalab} and whose final results are published at \citet{barriere-etal-2022-wassa}. The affect type of the task is emotion. The genre of the dataset is news articles, the modality is text, and the language is English.

The primary task for this project is the first subtask of the \citet{wassa-2022-codalab} shared task, Empathy Prediction, which consists of predicting both the empathy concern and the personal distress at the essay-level. This is a regression task. The dataset used in this project is the same as the one used in the shared task, and can be downloaded from \citet{wassa-2022-codalab}. The dataset contains empathic essay reactions to news stories, with associated Batson empathic concern and personal distress scores for each response. In addition to these scores, each response in the dataset contains gold standard labels for emotion, demographic information (age, gender, education, race, income) of the person who submitted the response, as well as the personality type of the writer.

The training data for this task consists of 1860 responses with gold standards for Empathy Prediction subtask. The development data consists of 270 responses with gold standard labels, and the test data contains 525 responses, but without gold standard labels.

The evaluation criteria for the Empathy Prediction task is the average Pearson correlation of the empathy scores and the distress scores. The evaluation for the test set requires predicting the outputs of the test set and submitting to the \citet{wassa-2022-codalab} website. The test scores are then generated on the CodaLab platform and are available for download.

\subsection{Adaptation Task} \label{section_secondary_task_description}

The adaptation task for this project is based on the WASSA 2023 Shared Task on Empathy Emotion and Personality Detection in Interactions \citep{wassa-2023-codalab}. This shared task builds on the shared task from \citet{wassa-2022-codalab} and includes dyadic (two person) text conversations about news articles. The dataset, described in \citet{omitaomu2022empathic}, can be downloaded from the \citet{wassa-2023-codalab} website. This dataset complements the Empathic Reactions dataset by \citet{buechel2018modeling} by providing conversational interactions rather than only first-person statements. 

The selected adaptation task for this project is Empathy and Emotion Prediction in Conversations, which involves predicting the perceived empathy, emotion polarity and emotion intensity at the speech-turn-level in a conversation. This is a regression task. The affect type of this task is emotion, and the genre of the dataset is news articles. The modality is text, and the language is English. This adaptation task differs from the primary task in that the primary task focuses on first-person text while the adaptation task focuses on turn-by-turn conversations.  One potential application for this adaptation task is to develop and evaluate conversational AI agents, such as ChatGPT, that are capable of producing and processing empathetic responses in human-AI interactions.

The training data for the adaptation task consists of 792 conversations with gold values for empathy and distress. Each of these conversations is further organized at the turn-level with 8,776 turns and has gold standard values for empathy, emotion polarity, and emotion intensity. The dev set consists of 208 conversations which are further organized at turn-level with 2,400 turns. Just like the training dataset, the dev set has the corresponding gold values. The test set consists of 136 conversations which are further organized at turn-level with 1,425 turns. Unlike the training dataset and the dev dataset, the test set does not have the corresponding gold values.

The evaluation criteria for the Empathy and Emotion Prediction in Conversations task is the average of the three Pearson correlations: Pearson correlation of empathy, Pearson correlation of emotional polarity, and Pearson correlation of emotional intensity. The evaluation for the test set requires predicting the outputs of the test set and submitting to the \citet{wassa-2023-codalab} website. The test scores are then generated on the CodaLab platform and are available for download.

\section{System Overview}

\subsection{Dataset repository and usage details}
The datasets for the primary task are part of the WASSA 2022 Shared Task on Empathy and Emotion Classification\footnote{\url{https://codalab.lisn.upsaclay.fr/competitions/834\#learn\_the\_details-overview}}. The training, development, and test datasets can be downloaded from the WASSA 2022 dataset link\footnote{\url{https://codalab.lisn.upsaclay.fr/competitions/834\#learn\_the\_details-datasets}}. 
As part of the WASSA 2022 dataset usage guidelines, these datasets must only be used for scientific or research purposes and the paper \citet{barriere-etal-2022-wassa} must be cited.

The datasets for the adaptation task is part of the WASSA 2023 Shared Task on Empathy Emotion and Personality Detection in Interactions\footnote{\url{https://codalab.lisn.upsaclay.fr/competitions/11167\#learn\_the\_details-overview}}. The training, development, and test datasets can be downloaded from the WASSA 2023 dataset link\footnote{\url{https://codalab.lisn.upsaclay.fr/competitions/11167\#learn\_the\_details-datasets}}. As part of the WASSA 2023 terms and conditions, these datasets should only be used for scientific or research purposes. Any other use is explicitly prohibited. Any use of the datasets must be accompanied with a citation of the associated paper \citep{omitaomu2022empathic}.

\subsection{Data exploration} \label{section_data_exploration}
\begin{figure}[h]
\centering
\includegraphics[height=5.8cm, width=10.5cm, keepaspectratio]{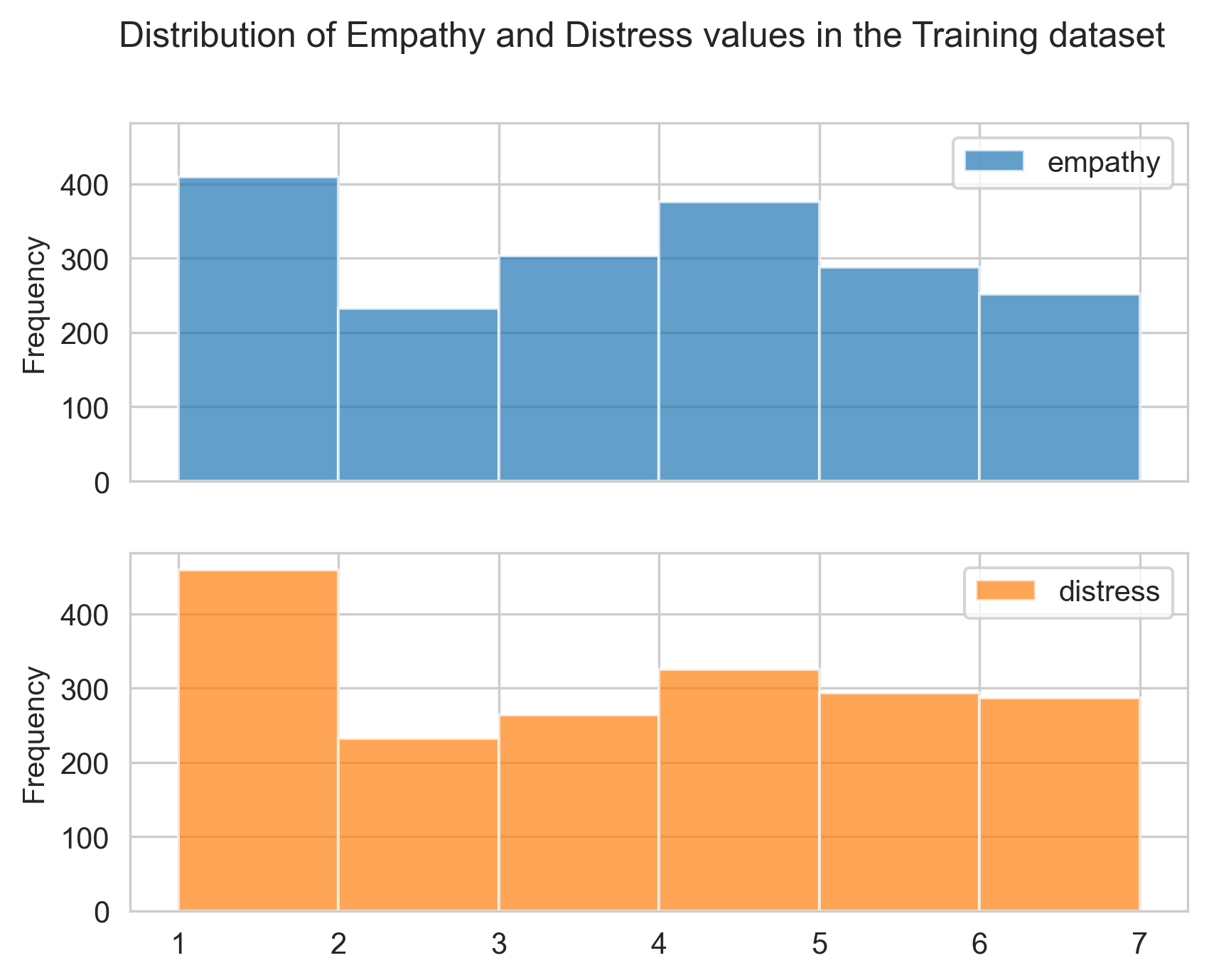}
\caption{Distribution of Empathy and Distress values in the training dataset for the primary task, indicating an imbalance in the distribution of samples}
\label{Empathy_Distress_distribution_train_dataset}
\end{figure}

For the primary task, the training dataset is comprised of 1860 rows, each of which containing three columns for empathy, distress and the essay.  The Dev dataset contains 270 rows with the same three columns. The test dataset contains 525 rows, but without the golden values. The distribution of the training dataset is shown in Figure~\ref{Empathy_Distress_distribution_train_dataset}. We observe that  the empathy and distress values in the training and dev datasets are imbalanced, with a higher concentration of density between values 1 and 2.

\begin{figure}[h]
\centering
\includegraphics[height=5.8cm, width=10.5cm, keepaspectratio]{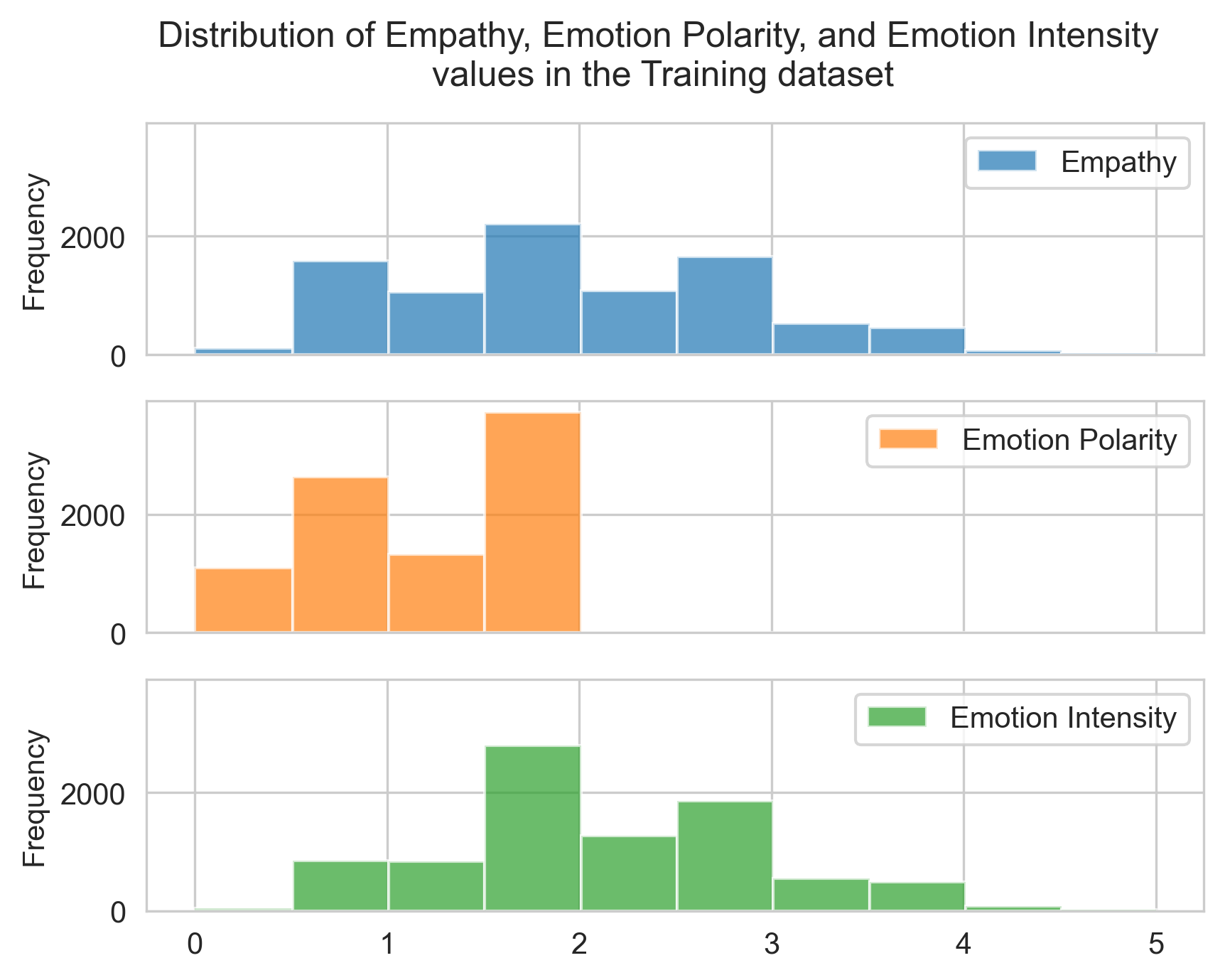}
\caption{Distribution of Empathy, Emotion Polarity, and Emotion Intensity values in the training dataset for the adaptation task, indicating an imbalance in the distribution of samples}
\label{adaptation_distribution_train_dataset}
\end{figure}

For the adaptation task, the training dataset consists of 792 conversations, each further organized into 8,776 turns. The dataset has the corresponding values for empathy, emotion polarity, and emotional intensity. The dev set consists of 208 conversations, each further organized into 2,400 turns, with the same three target values of empathy, emotion polarity, and emotion intensity. The test set consists of 136 conversations, each further organized into 1,425 turns. The distribution of the target features in the training set is shown in Figure~\ref{adaptation_distribution_train_dataset}. We observe that the empathy and emotion polarity values are between 0 and 5, while the values for emotion polarity ranges from 0 to 2. Additionally, we observe an imbalance in the distribution of data values. For empathy and emotion polarity, there are fewer samples near 0 and 5 than towards other numbers. Similarly, for emotion polarity, the distribution of samples is unbalanced, with more samples near 1 and 2 than near 0.

\subsection{Architecture overview}

The architecture diagram in Figure~\ref{architecture-overview} shows an overview of the system. The first module in this system performs data loading, data exploration, and preprocessing. The training and development datasets are loaded into pandas dataframes. The golden values for the dev dataset are joined with the dev instances to facilitate comparison evaluation. The observations from data exploration are described in the previous section. From a preprocessing perspective, the text from the essays are encoded using BPE tokenization before calling the Azure OpenAI embedding model. The other embedding models do not need preprocessing and are therefore kept as is. The essay text was the only feature used for the initial system during the first stage. The texts in the essay have been converted to dense vectors using the embedding models described in Section \ref{section_embedding}.

During the second stage, three revisions have been made to this architecture. In the first revision, the Model Selection and Model Training portions of the architecture have been enhanced. The updates include updating the dropout method, adding more modern activation function, and finetuning the training process. These updates are further detailed in Section \ref{hyperparameter_tuning}. The second revision focuses on addressing class imbalance using Stratified Sampling. This revision is further detailed in Section \ref{section_sds}. The third revision focuses on the Feature Extraction portion of the architecture. Four lexical resources have been used to expand the number of features that are used during training. This revision is detailed in Section \ref{section_lexicon}.

During the final stage, the final end-to-end system has been prepared. The revision includes the use of ensembles and this revision is detailed in Section \ref{section_primary_final_approach}. The affect recognition system finalized for the primary task has been adapted for the adaptation task. The approach for the adaptation is detailed in Section \ref{section_adaptation_approach}.

\subsection{The hardware} 

The embeddings for sentence-transformer models were initially generated on CPU, but this was found to be very time consuming. Subsequent embeddings were generated on a NVIDIA Tesla T4 GPU hosted on Google Colab. The embeddings for the text-embedding-ada-002 model were generated using Azure OpenAI API. The values of the embedding vector are stored in a data store to allow efficient modeling. This NVIDIA Tesla T4 GPU-based hardware has been used to train the Neural Network models.

\begin{figure*}[!hbt]
\centering
\includegraphics[width=\textwidth]{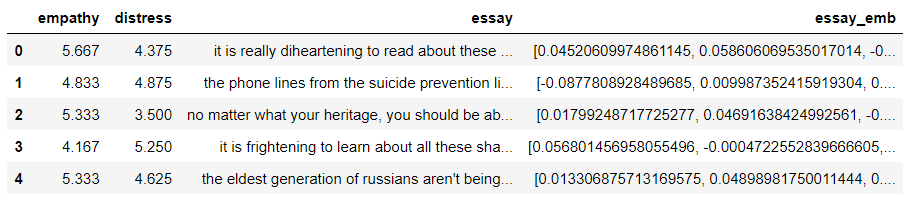}
\caption{Samples from the Training Dataset with Embeddings (Sentence Transformer)}
\label{Training Dataset with Embedding SentenceTransformer}
\end{figure*}

\section{Approach} \label{section_approach}

\subsection{Initial system}

For the initial system presented in the first stage, a Feed-Forward Neural Network has been implemented. The PyTorch library has been used to create the Neural Network models. The implementation is based on the FFN architecture proposed in \citet{buechel2018modeling} with two hidden layers (256 and 128 units, respectively) with ReLU activation. The sentence-level embeddings are used as features for this model. Dropout layers with p=0.5 values have been added before every linear layer to help reduce overfitting. 20\% of the training set is set aside to act as a validation set. MSE loss function has been used as the loss for the training and AdamW with learning rate of 1e-4 has been used as the optimizer. The seed value has been set for numpy and pytorch to help with reproducibility of the results. The validation set is used to select the model with the lowest MSE when running the training loop for 100 epochs. The model weights have been saved so that these weights can be used during the evaluation and scoring steps of the project.

\subsection{Embedding models} \label{section_embedding}

For the initial system presented in the first stage, we have used four different embedding models.

The all-MiniLM-L6-v2 is a sentence-transformer model \citep{wang2020minilm}. This model maps sentences and paragraphs to 384 dimensional dense vector space which captures semantic information. By default, input text longer than 256 word pieces is truncated. The MiniLM is a six layer version of MiniLM model created by Microsoft \citep{wang2020minilm}. Figure~\ref{Training Dataset with Embedding SentenceTransformer} shows a snippet of the training dataset with a few values of the sentence-transformer embedding.

The all-mpnet-base-v2 is a sentence-transformer model that maps sentences and paragraphs to a 768 dimensional dense vector space. By default, input text longer than 384 word pieces is truncated. This model is based on the MPNet model created by Microsoft \citep{song2020mpnet}.

The all-roberta-large-v1 is a sentence-transformer model that maps sentences and paragraphs to 1024 dimensional dense vector space. By default, input text longer than 128 word pieces is truncated. This model is based on RoBERTa developed by the University of Washington and Facebook AI \citep{liu2019roberta}.

The text-embedding-ada-002 is an embedding model created by OpenAI and served from Microsoft Azure \citep{neelakantan2022text} \citep{azureopenai}. This model maps a list of tokens to a dense vector of 1536 dimensions and replaces five separate models for text search, text similarity, and code search tasks. This model uses cl100k\_base tokenizer that uses BPE tokenization and has a limit of 8191 maximum tokens.

\begin{figure}[!hbt]
\centering
\subfloat[]{\label{subfig:10}\includegraphics[height=5.8cm, width=10.5cm, keepaspectratio]{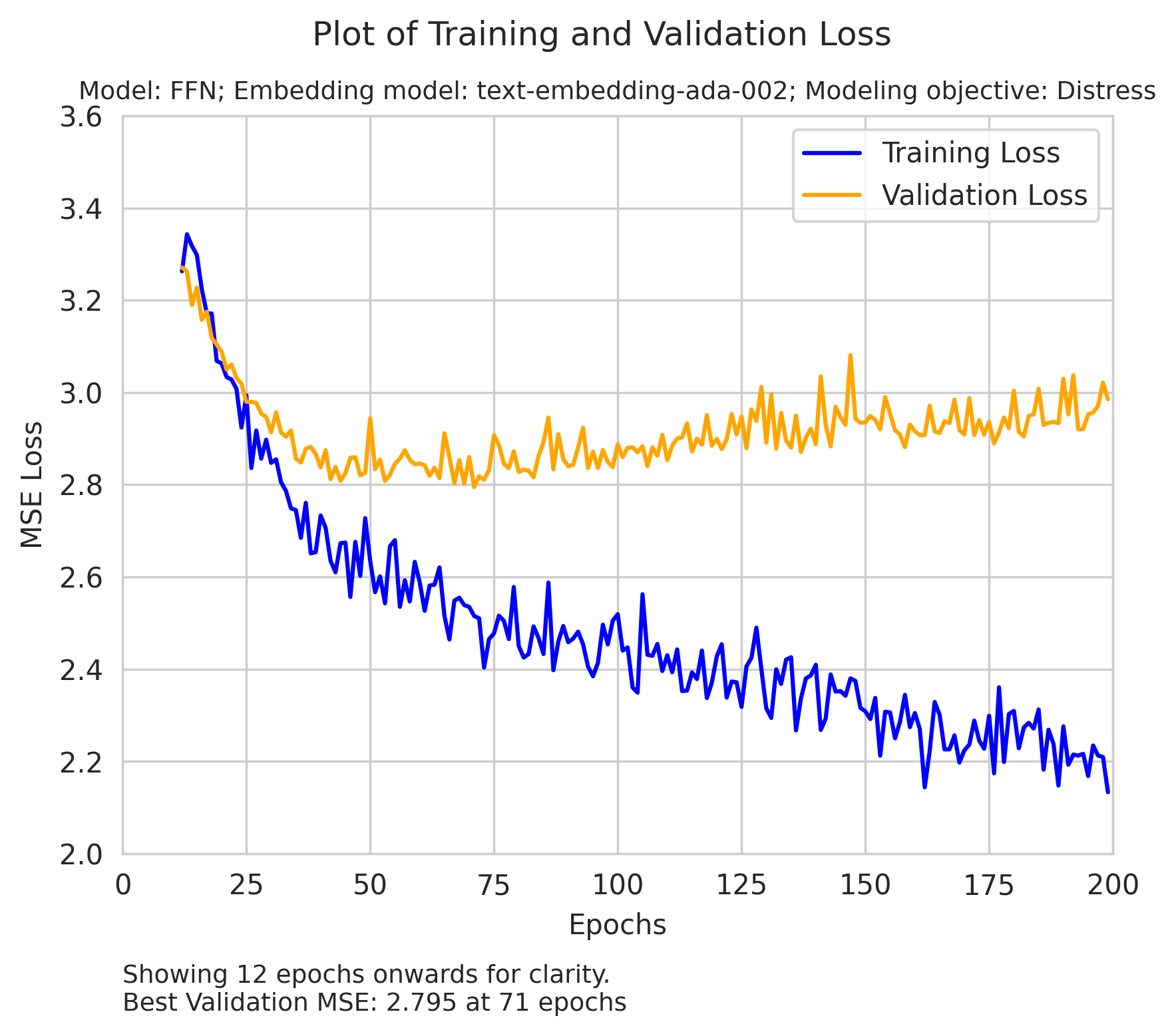}} \\
\subfloat[]{\label{subfig:11}\includegraphics[height=5.8cm, width=10.5cm, keepaspectratio]{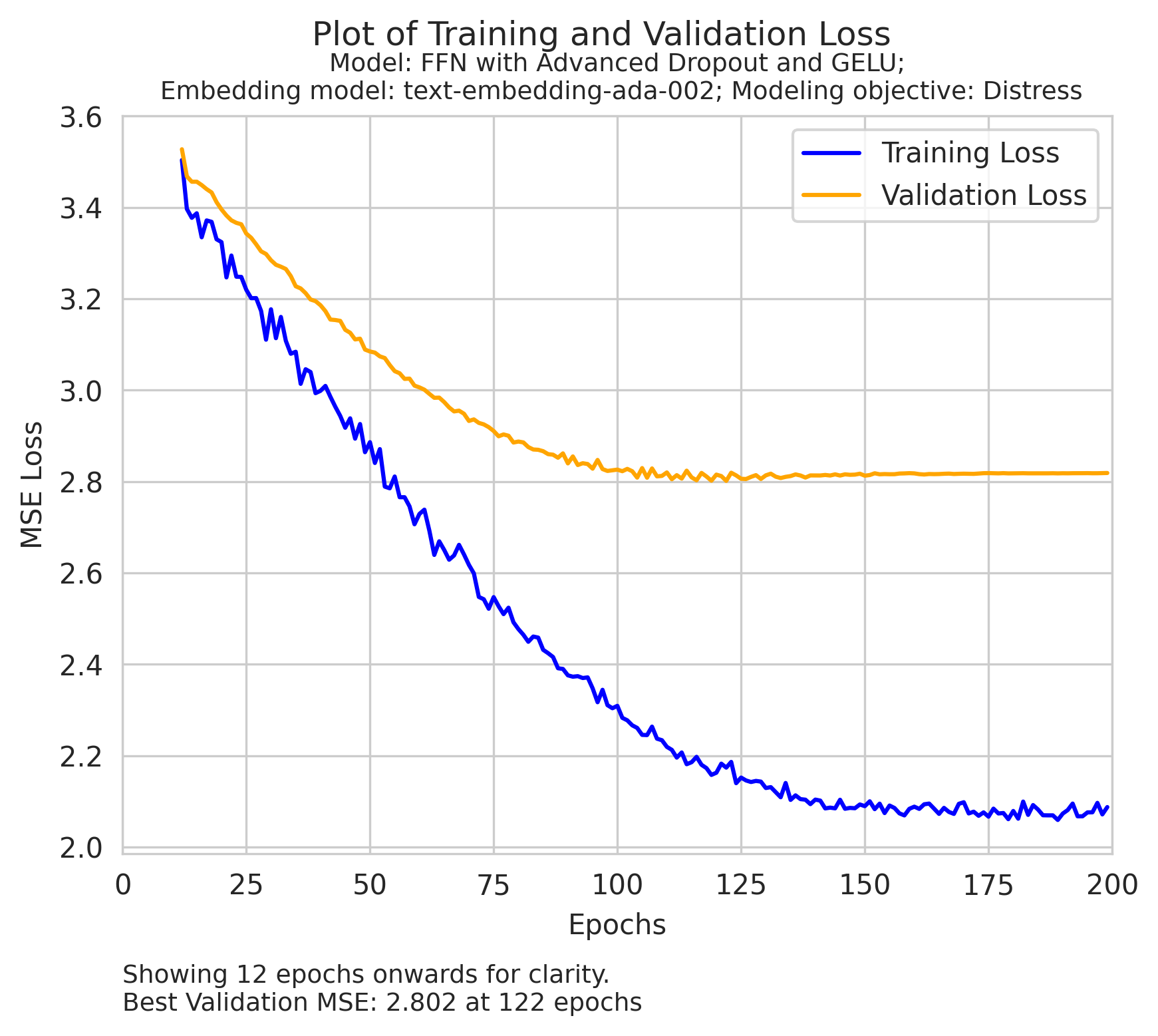}}
\caption{Training and validation losses:      \protect\subref{subfig:1} Before hyperparameter tuning \protect\subref{subfig:2} After hyperparameter tuning}
\label{plots_hyperparameter_before_after}
\end{figure}

\subsection{D\#3 Primary Task Revision \#1: Hyperparameter tuning} \label{hyperparameter_tuning}

Previously during the first stage of this project, a Feed Forward Network was used with two hidden layers (of 256 and 128 neurons), RELU activation functions, and dropout layers with p=0.5. During the second stage multiple experiments were performed to finetune this model architecture.

The first set of experiments focused on the dropout layers and the activation function. As mentioned in the discussion section of the first stage, the model tends to overfit the training dataset. One of the methods of mitigating overfitting is to use dropout. However, getting the values of the dropout rate (p) right requires hyperparameter tuning. Such tuning is generally very time consuming, requiring retraining the entire model with each option of the dropout rate. In an alternate approach, \citet{Xie_2021} proposed an advanced dropout technique that adaptively adjusts the dropout rate, resulting in a stable convergence of dropout rate and superior ability of preventing overfitting.

The paper \citet{hendrycks2020gaussian} proposes a novel activation function, called Gaussian Error Linear Unit (GELU). This nonlinearity weights inputs by their value, rather than gates inputs by their sign as in ReLUs. Experiments in this paper indicates that GELU exceeds the accuracy of ReLU consistently, and therefore we have used GELU in the updated model architecture.

In addition to these updates to the model, a PyTorch Learning Rate Scheduler has been implemented during model training. This scheduler has been configured to reduce the learning rate when reaching a plateau, with a factor of 0.8 and a patience value of 3. The initial learning rate of the AdamW optimizer has been kept the same as during the first stage to allow for relative comparisons.

The results of these changes to the model architecture are shown in Table \ref{results-table-D3} and the associated plots of train and validation losses are show in Figure~\ref{plots_hyperparameter_before_after}. The results of the Pearson correlation of empathy scores and distress scores are comparable to the best performing model from the first stage. We can observe from the plots that overfitting has been mitigated from this model and we can further assert that even when the model is trained for double the number of epochs (200 epochs compared to 100 epochs from the first stage) the model does not overfit. These changes to the architecture can allow for longer training resulting in a more stable training process.

\begin{figure}[h]
\centering
\includegraphics[height=5.8cm, width=10.5cm, keepaspectratio]{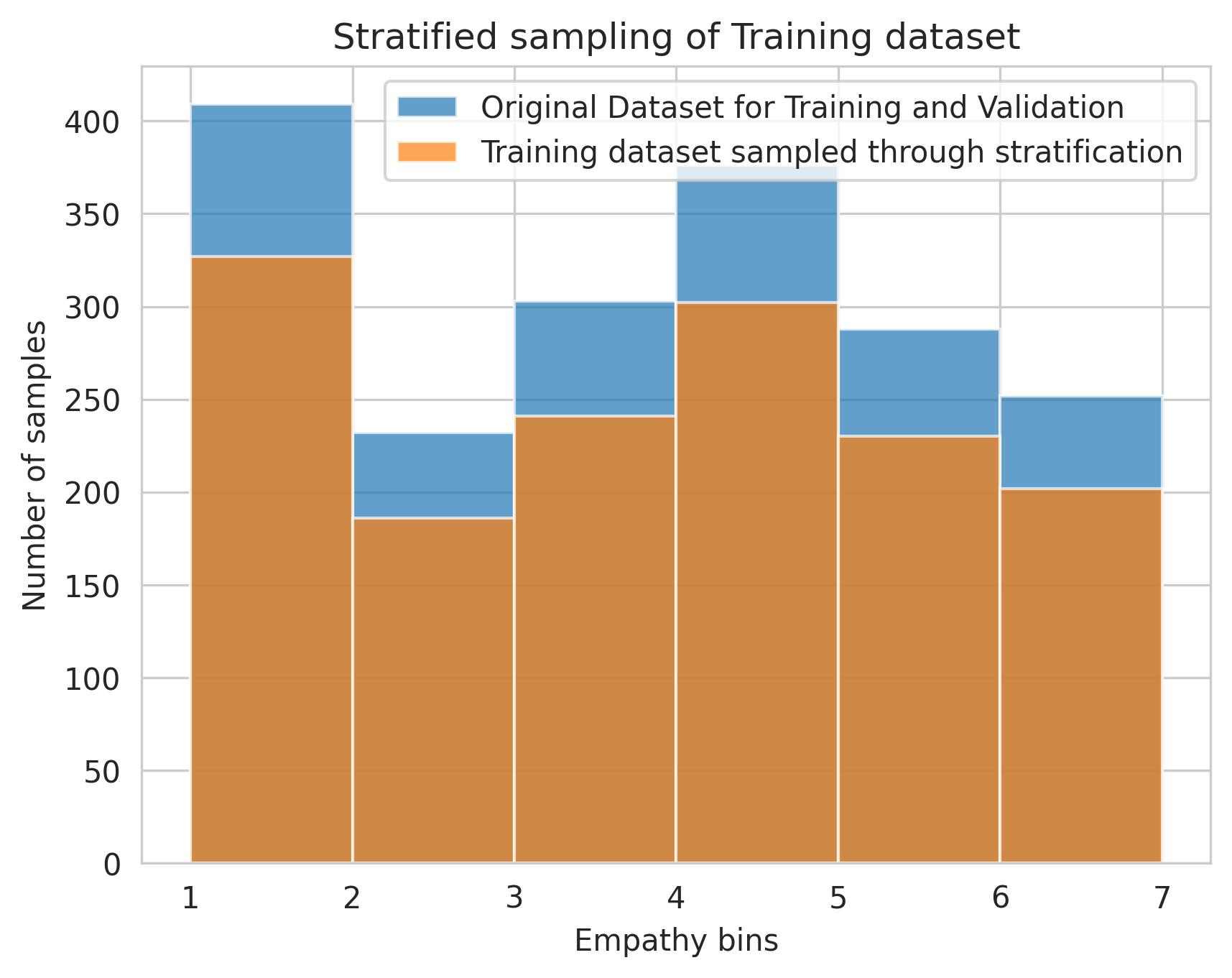}
\caption{Distribution of the 80\% Training dataset sampled from the original WASSA Dataset}
\label{data_sampling}
\end{figure}

\subsection{D\#3 Primary Task Revision \#2: Handling class imbalance using Stratified Data Sampling} \label{section_sds}

Stratified Sampling approach has been used to address the class imbalance. This method of sampling preserves the same distribution of each target class in the training and validation sets as in the original dataset. This approach was implemented using Scikit-learn's \texttt{stratify} parameter in train\_test\_split method. The Figure~\ref{data_sampling} shows the distribution of the original dataset and the distribution of the training dataset after stratification. We can observe that the same proportion of the original dataset has been retained in the training dataset. Once the model is trained with this training dataset, we can observe that the Pearson correlation for the empathy score and distress score increases by $9.4\%$ and $5.4\%$ respectively over the results from Revision \#1. These results show that stratified sampling has a significant impact to the performance scores. The results are updated in Table \ref{results-table-D3}.

\begin{table*}
\centering
\begin{tabular}{lccc}
\hline & \textbf{Empathy} & \textbf{Distress} & \textbf{Mean} \\ \hline

FNN baseline & .379 & .401 & .390\\
FNN with \texttt{all-MiniLM-L6-v2} embedding & .379 & .370 & .375\\
FNN with \texttt{all-mpnet-base-v2} embedding & .386 & .324 & .355\\
FNN with \texttt{all-roberta-large-v1} embedding & .395 & .360 & .378\\
FNN with \texttt{text-embedding-ada-002} embedding & .438 & .426 & .432\\

\hline
\end{tabular}
\caption{\label{results-table-D2} Table of results for D\#2. Model performance for predicting empathy and distress in Pearson correlations, with row-wise mean. Higher scores correspond to stronger correlations.}
\end{table*}

% Results after advanced dropout and Gelu and data sampling
\begin{table*}
\centering
\begin{tabular}{lccc}
\hline & \textbf{Empathy} & \textbf{Distress} & \textbf{Mean} \\ \hline

FNN baseline & .379 & .401 & .390\\
FNN best performing model from D\#2 & .438 & .426 & .432\\
D\#3 Revision \#1: FNN, Advanced Dropout, GELU & .434 & .425 & .430\\
D\#3 Revision \#2: FNN, Advanced Dropout, GELU, and SDS & .475 & .448 & .462\\
D\#3 Revision \#3: FNN, Advanced Dropout, GELU, Lexicons, and SDS & .481 & .456 & .469\\

\hline
\end{tabular}
\caption{\label{results-table-D3} Table of results for D\#3: Model performance for predicting empathy and distress in Pearson correlations, with row-wise mean. Higher scores correspond to stronger correlations. The revisions include Advanced Dropout, GELU, Lexicons, and Stratified Data Sampling (SDS). All results beyond the baseline are with the \texttt{text-embedding-ada-002} embedding model}
\end{table*}

% table for the final primary task %
\begin{table*}
\centering
\begin{tabular}{lccc}
\hline & \textbf{Empathy} & \textbf{Distress} & \textbf{Mean} \\ \hline

FNN baseline & .379 & .401 & .390\\
FNN best performing model from D\#2 & .438 & .426 & .432\\
FNN best performing model from D\#3 & .481 & .456 & .469\\
Final end-to-end model: DevTest results & .488 & .471 & .480\\
Final end-to-end model: EvalTest results & .520 & .521 & .521\\

\hline
\end{tabular}
\caption{\label{results-table-D4-primary} Table of results for D\#4 Primary Task: Model performance for predicting empathy and distress in Pearson correlations, with row-wise mean. Higher scores correspond to stronger correlations.}
\end{table*}

% table for the adaptation task %
\begin{table*}
\centering
\begin{tabular}{l c c c c}
\hline & \textbf{Empathy} & \textbf{Emotional Polarity} & \textbf{Emotional Intensity} & \textbf{Mean} \\ \hline
Baseline Results & .575 & .339 & .369 & .428\\
Adaptation Task: DevTest Results & .701 & .775 & .756 & .744\\
Adaptation Task: EvalTest Results & .761 & .652 & .691 & .702\\
\hline
\end{tabular}
\caption{\label{results-table-D4-adaptation} Table of results for D\#4 Adaptation Task: Model performance for predicting Empathy, Emotional Polarity, and Emotional Intensity in Pearson correlations, with row-wise mean. Higher scores correspond to stronger correlations.}
\end{table*}

\subsection{D\#3 Primary Task Revision \#3: Lexicon features} \label{section_lexicon}

For the revised system presented in the second stage, we have created 48 additional features based on 4 different word-level lexicons. Essays were preprocessed via NLTK’s tokenizer and WordNet’s lemmatizer before applying the corresponding lexicons.

The NRC Word-emotion association lexicon \citep{Mohammad13}  provides markers on word relations to the eight basic emotions of Plutchik’s wheel of emotions, as well as association to general polarity (i.e. positive and negative feelings). The lexicon was built on the union of three datasets (Google top n-grams list, WordNet Affect Lexicon, and General Inquirer), totaling 14,154 words. Annotation for emotions was carried out through crowdsourcing via Mechanical Turk requests. Features derived from this lexicon include word counts by essay for each emotion and their corresponding ratio (normalized by essay length).

The MPQA subjectivity lexicon \citep{wilson-etal-2005} contains scores for prior polarity (i.e. positive, negative, neutral, both) as well as contextual information (i.e. whether the examined word has strong or weak subjectivity) and part-of-speech information over 6,886 words. The lexicon was constructed by combining existing subjectivity corpora (e.g. Multi-perspective Question Answering Opinion corpus) with additional dictionaries and thesaurus, as well as positive/negative word lists from the General Inquirer. Features generated from this lexicon included word counts and ratios for type of subjectivity, part-of-speech tags and polarity.

The NRC VAD lexicon \citep{vad-acl2018} consists of 19,852 words which have been annotated for valence, arousal and dominance scores. The lexicon’s dataset is comprised of terms from various sources including the NRC Emotion lexicon, General Inquirer, ANEW, amongst others. Annotation was carried out through a Best-Worst scaling approach, which asked annotators crowdsourced via the CrowdFlower platform to rank tuples of 4 words from least to most valence, arousal or dominance. Features created from this lexicon include the mean of the valence, arousal and dominance scores for each word in the essay.

The verbal polarity shifters lexicon \citep{schulder-etal-2018-introducing} contains annotations for words which can cause a shift in polarity (i.e. positive or negative feeling) or not. The lexicon consists of 10,577 lemmas sourced from WordNet (a lexicon at the lemma-sense level is also available but was not employed for this model). The annotation process was performed by an expert with experience in both linguistics and annotation, and a second annotator re-annotated 400 word senses for validation. Features derived from this lexicon include the count of shifter words that appear in the essay.

\subsection{D\#4 Primary Task Final end-to-end system: Ensemble methods} \label{section_primary_final_approach}

The primary task system for the final stage is a small ensemble consisting of our model from the second stage and two Support Vector Regression (SVR) models from the \texttt{svm} module of Scikit-learn \citep{scikit-learn}. Each ensemble prediction is the average of the predictions of the three models. The prediction of each model receives equal weight in the calculation of the ensemble's prediction.

Apart from the models' kernels, both SVRs use the default Scikit-learn \texttt{svm.SVR} parameters, including epsilon equal to 0.1 and the regularization parameter C equal to 1.0. One SVR uses a 3rd-degree polynomial kernel with the independent term (\texttt{coef0}) set to 0.0, the other uses an RBF kernel. The SVRs are trained on the entire \citet{wassa-2022-codalab} training set. Sample weight during training is determined using a linear function of the distance of each sample's true value from the midpoint of the 1-7 empathy or distress scale. The weight of sample $s$ is computed as $w_s = | m - g_s | + 1$, where $m$ is the midpoint of the scale for the relevant emotion and $g_s$ is the true value for sample $s$.

A major motivating factor for our ensemble's final configuration was avoiding overfitting. To avoid overfitting of model weights within the ensemble to a validation set, models were assigned equal weight for the ensemble's predictions. Following similar reasoning, SVR hyperparameters were not tuned to avoid overfitting the models' weights to a validation set. An added benefit of not using a validation set during SVR training or model weight selection within the ensemble was that the SVRs were able to be trained on the entire \citet{wassa-2022-codalab} training set.

Two considerations for selecting ensemble models were informed by the Scikit-learn User Guide for the \texttt{ensemble} module's \texttt{VotingRegressor} class\footnote{\url{https://scikit-learn.org/stable/modules/ensemble.html\#voting-regressor}}. One of these considerations was that models were selected for roughly similar performance on the dev data as our second stage model. SVR models were found to meet this requirement. The other consideration was that the approach of the other models selected for the ensemble would complement our second stage model. 

Error analysis of our second stage model in Section \ref{section_primary_error_analysis} revealed that the model's performance was more consistent on data samples with true values near the midpoint of the empathy or distress scale. Error was higher on data samples with true values near the ends of the scale. To promote better ensemble performance on these samples, they were given higher weight during SVR training.

\subsection{D\#4 Adaptation Task} \label{section_adaptation_approach}

The adaptation task follows a similar architecture as the primary task. The datasets from \citet{wassa-2023-codalab} are first loaded into pandas DataFrames for exploration. During the feature extraction stage, texts from essays and text from turn-level conversations are selected. These texts are encoded using the byte-pair-encoding subword tokenization and text embeddings are obtained using the \texttt{text-embedding-ada-002} embedding model, as described in Section \ref{section_embedding}. Each of the texts is pre-processed to create the lexicon features as described in Section \ref{section_lexicon}. These lexicon features are standardized and normalized before concatenating with the features obtained from the embedding model. 

Since the adaptation task uses turn-level conversations, the feature vectors are adapted to accommodate this task. The premise applied is that the affect of the current turn-level text depends on three factors: the affect of the essay written by a person, the affect of the text at the current turn-level, and the affect of all the previous turn-level conversations for both the persons involved in the conversation. To obtain the feature vector of all the previous turn-level conversations, a new vector is created using the centroid of all the previous turn-level conversations. The three  vectors (for the essay, the current turn-level conversation, and the previous turn-level conversations for both persons) are concatenated to produce the final feature vector for modeling. This approach is used to save the modeling DataFrames for train, dev, and test datasets.

\begin{figure}[!hbt]
\centering
\subfloat[]{\label{subfig:12}\includegraphics[height=5.8cm, width=10.5cm, keepaspectratio]{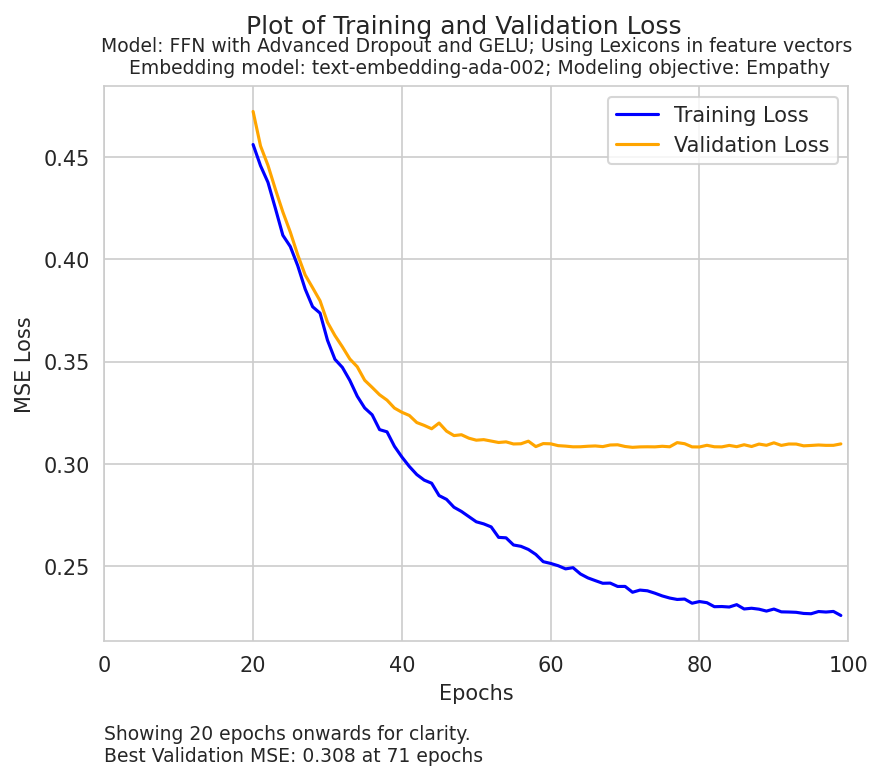}} \\
\subfloat[]{\label{subfig:13}\includegraphics[height=5.8cm, width=10.5cm, keepaspectratio]{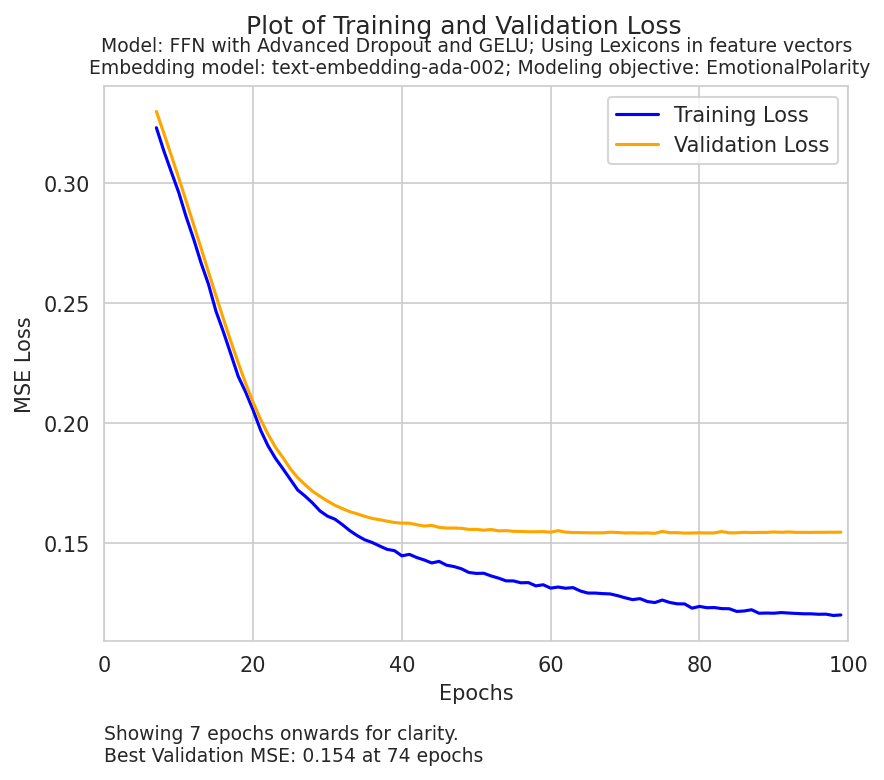}} \\
\subfloat[]{\label{subfig:14}\includegraphics[height=5.8cm, width=10.5cm, keepaspectratio]{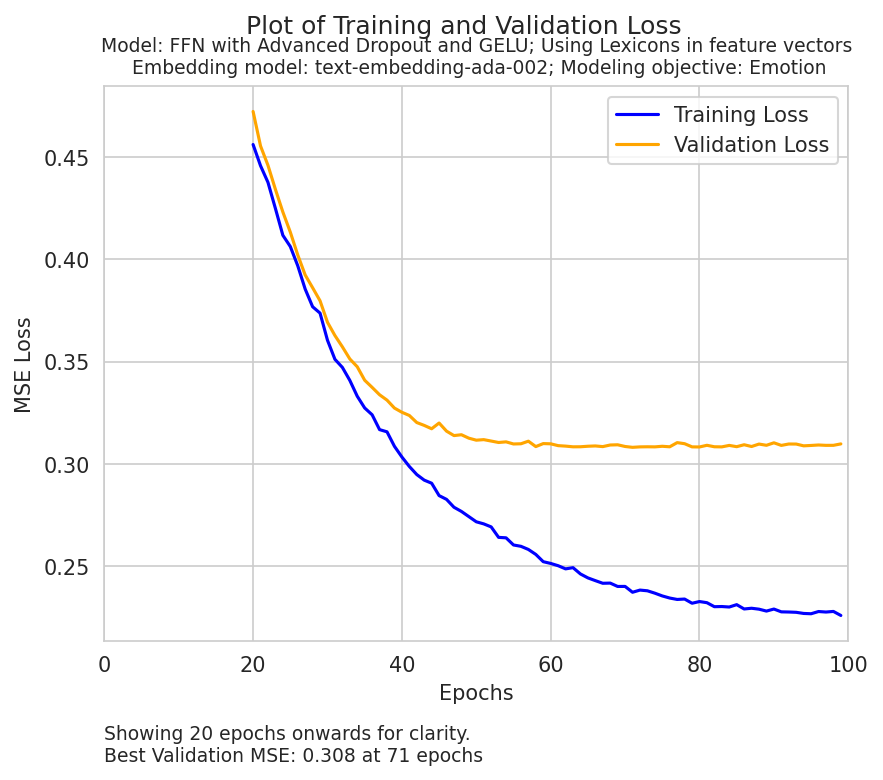}}
\caption{Training and validation losses for the adaptation task indicating a stable training process without overfitting:      \protect\subref{subfig:1} Empathy \protect\subref{subfig:2} Emotional Polarity \protect\subref{subfig:3} Emotional Intensity}
\label{plots_adaptation}
\end{figure}

The structure of the feed-forward neural network model that was used in the primary task has been reused for the adaptation task. This FFN model uses Advanced Dropout and GELU activation function as described in Section \ref{hyperparameter_tuning}. The FFN model has two hidden layers of 256 and 128 neurons. The complete training set is used for model training, therefore stratified sampling is not required. The loss function is the mean-squared-error, and AdamW is used as the optimizer. Learning rate scheduler has been used with a factor of 0.8 and patience of 3. The initial learning rate is set as 2e-5 for the models for Emotional Polarity and Emotional Intensity. The initial learning rate is set at 1e-5 for modeling Empathy. The minimum learning rate has been set at 1e-6 in the learning rate scheduler. Each of the models is trained for 100 epochs. Figure~\ref{plots_adaptation} shows the plots for the training and validation losses for each of the three models. We can observe that the training and validation curves do not indicate overfitting. This stable training process is attributed to the various hyper-parameter tuning approaches, as described in Section \ref{hyperparameter_tuning}, that have been adapted from the primary task.

\section{Result} \label{section_results}

\subsection{Initial system results}
Model performance for predicting empathy and distress is reported in terms of Pearson correlation, including row-wise mean for the empathy and distress scores. Results are presented in Table~\ref{results-table-D2}, with the first row representing the FNN baseline based on the \citet{buechel2018modeling} model. All subsequent reported FNN's follow the same base network architecture, but vary the models used to generate the input embeddings. Similar correlations are observed for all models, with the exception of the \texttt{text-embedding-ada-002} model which achieved much higher scores. For the \texttt{text-embedding-ada-002} model, we observe that the Pearson scores for Empathy increases by 15.57\% and the scores for Distress increases by 6.23\%.

\begin{figure*}
\centering
\subfloat[]{\label{subfig:10}\includegraphics[keepaspectratio]{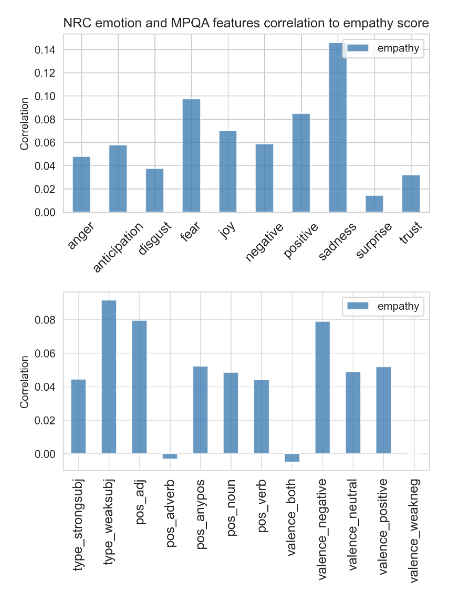}}
\subfloat[]{\label{subfig:11}\includegraphics[keepaspectratio]{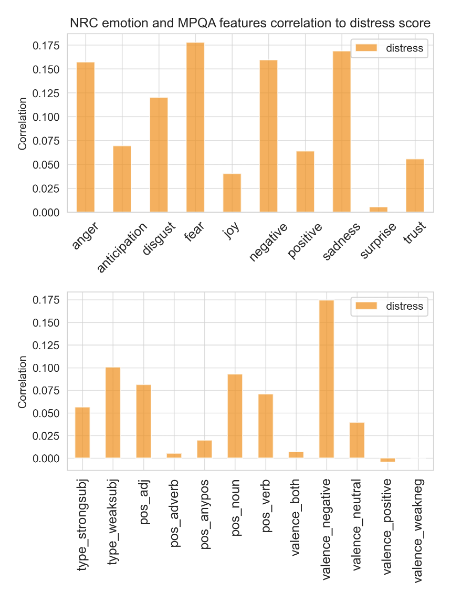}}
\caption{Correlation of lexicon features with:      \protect\subref{subfig:1} empathy scores \protect\subref{subfig:2} distress scores}
\label{correlation_lexicon_features}
\end{figure*}

\subsection{Primary Task D\#3 Revised system results} \label{section_revised_results}
The results from the revised system are presented in Table~\ref{results-table-D3}. The first row presents the FNN baseline based on the \citet{buechel2018modeling} model, and the second row presents the evaluation scores for the best performing model from the first stage, which was the FNN with \texttt{text-embedding-ada-002} embedding. Each of the following rows presents the evaluation scores after making incremental revisions to the system.

The revision \#1 uses the FNN model with the addition of Advanced Dropout layers and GELU activation units as described in the Section \ref{hyperparameter_tuning}. This revision results in an increase of 10.3\% in the mean score from the baseline and maintains the mean score at a similar level as from the initial system from D\#2. The Revision \#2 incrementally applies Stratified Data Sampling as described in Section \ref{section_sds}. This revision leads to an 18.5\% increase in the mean scores from the baseline, and 6.9\% increase compared to the scores from the D\#2 initial system. The Revision \#3 incrementally applies lexical features as described in Section \ref{section_lexicon}. This revision leads to an 20.3\% increase in the mean scores from the baseline, and 8.6\% increase compared to the scores from the first stage. We can observe that each of the revisions incrementally builds on the previous work, resulting in an overall model that significantly outperforms both the baseline and the model from the first stage.

\subsection{Primary Task Final end-to-end system results} \label{section_primary_final_results}

Results for our final primary task system on the dev (DevTest) and test (EvalTest) sets are shown in Table~\ref{results-table-D4-primary}. On the dev set, our final system's mean Pearson correlation is a 2.3\% improvment from our best second stage model and a 23.1\% improvement over our baseline. Our final system scores higher on the test set than on our dev set. The mean Pearson correlation on the test set is 0.521, indicating that our system's predictions are strongly positively correlated with the true values of the test set. Our final system for the primary task modestly improves our second stage dev scores and performs well on the test data.

\subsection{D\#4 Adaptation Task Results} \label{section_adaptation_results}

The baseline scores for the adaptation task are obtained from \citet{omitaomu2022empathic}. The authors of this paper use a Gated Bi-RNN with attention layer for the baseline model. This baseline model uses fast-text embeddings of 300-dimensions. For this baseline model, the authors used 5,821 conversation turns and split the data randomly into 70\%/15\%/15\% for train/dev/test, respectively. Please note that this split is different from the one used in the \citet{wassa-2023-codalab} shared task. The shared task includes specific train/dev/test datasets as detailed in Section \ref{section_data_exploration}. 

The results from the adaptation task are presented in Table~\ref{results-table-D4-adaptation}. The model from the adaptation task exhibits an increase of 73.83\% in the mean score from the baseline when evaluated on the dev dataset (DevTest Results). When evaluated on the test dataset (EvalTest Results), this model leads to an increase of 64.02\% in the mean score from the baseline. We can observe that the model implemented in the adaptation task substantially exceeds the scores from the baseline.

\subsection{Error analysis for the primary task} \label{section_primary_error_analysis}

The error analysis is similar to the one done in the paper \citet{kuijper-etal-2018-ug18}. We performed a manual error analysis on the data showing the maximum deviation as well as overall positive and negative deviation. Positive deviation is when the value of gold standard is higher than the predicted value. Negative deviation is when the value of the gold standard data is less than predicated value.

\begin{figure*}[ht]
\centering
\includegraphics[width=\textwidth]
{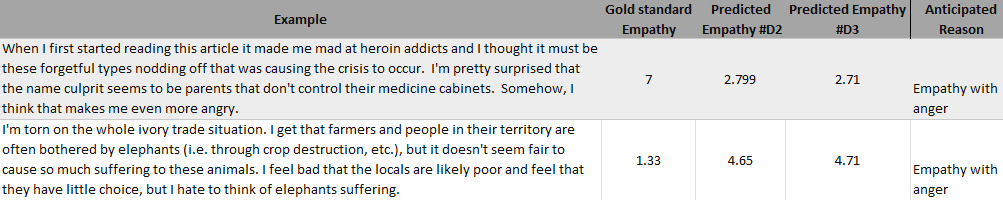}
\caption{Empathy data example for maximum deviation from gold standard}
\label{empathy_error}
\end{figure*}

\begin{figure*}[!ht]
\centering
\includegraphics[width=\textwidth]{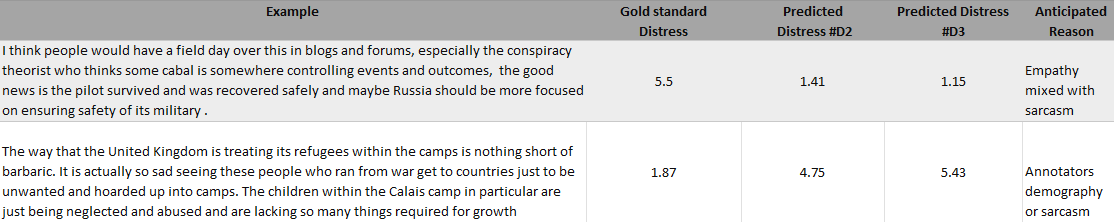}
\caption{Distress data example for maximum deviation from gold standard}
\label{distress_error}
\end{figure*}

For D\#2 the average positive deviation for empathy is 1.5 and for distress is 1.54. The average negative deviation for empathy is -1.33 and for distress is -1.41. Similarly, for D\#3 the average positive deviation for empathy is 1.33 and for distress is 1.47. The average negative deviation for empathy is -1.40 and for distress is -1.41. So, we can observe that the average of both empathy and distress remains almost the same. We present the following key takeaways and reasons behind these observations. First, the overall prediction given by the gold standard depends on the annotator demography, which may affect their perception of empathy/distress. Another reason is when text is depicting empathy mixed with anger then the model may produce a confused result and deviate from the gold standard value. The example shown in Figure {\ref{empathy_error}} and Figure {\ref{distress_error}} gives few data examples for empathy and distress deviations from the gold standard. 

\begin{table*}
\centering
\begin{tabular}{lrrrr}
\hline & \textbf{True values} & \textbf{D\#2 predictions} & \textbf{D\#3 predictions} & \textbf{D\#4 predictions}\\ \hline
Empathy mean & 3.558 & 3.412 & 3.603 & 3.453\\
Empathy standard deviation & 1.861 & 0.848 & 0.960 & 1.171\\
Distress mean & 3.773 & 3.707 & 3.722 & 3.628\\
Distress standard deviation & 1.932 & 1.055 & 1.157 & 1.353\\
\hline
\end{tabular}
\caption{\label{center and spread measures} Measures of center and spread for empathy and distress true values and D\#2, D\#3 and D\#4 predictions. }
\end{table*}

\begin{table*}
\centering
\begin{tabular}{lrrr}
\hline & \textbf{D\#2 absolute error} & \textbf{D\#3 absolute error} & \textbf{D\#4 absolute error} \\ \hline
True empathy distance from center & 0.540 & 0.550 & 0.331 \\
True distress distance from center & 0.527 & 0.468 & 0.352 \\
\hline
\end{tabular}
\caption{\label{centrality and error correlations} Pearson correlations of model prediction absolute error and distance of true label from the center of the 1-7 scale. }
\end{table*}

\begin{figure*}
\centering
\mbox{\subfloat[]
{\label{subfig:1}\includegraphics[width=0.45\textwidth]
{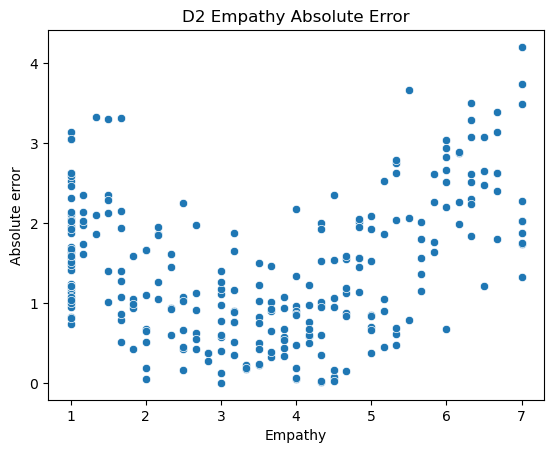}}}
\mbox{\subfloat[]
{\label{subfig:2}\includegraphics[width=0.45\textwidth]
{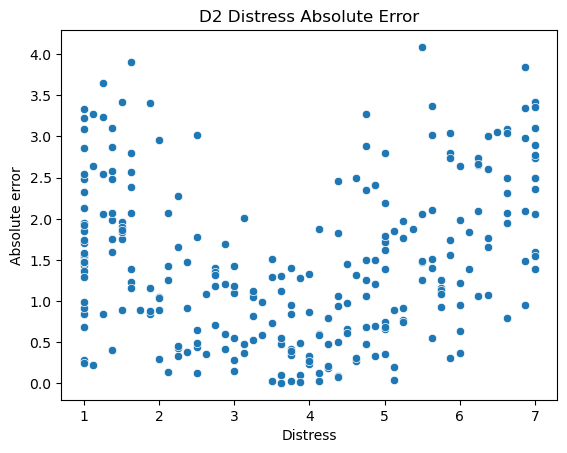}}}
\mbox{\subfloat[]
{\label{subfig:3}\includegraphics[width=0.45\textwidth]
{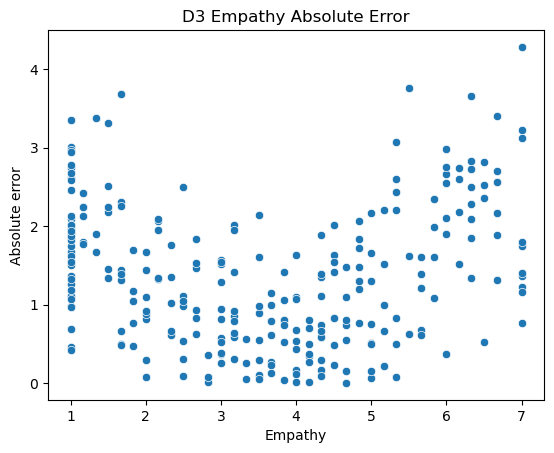}}}
\mbox{\subfloat[]
{\label{subfig:4}\includegraphics[width=0.45\textwidth]
{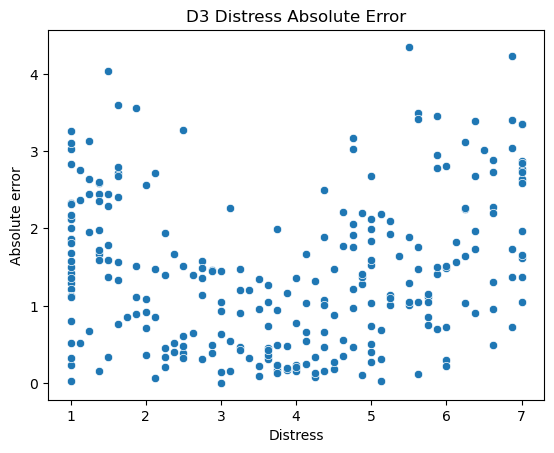}}}
\caption{Absolute error of predictions with respect to true values of empathy and distress: \protect\subref{subfig:1} D\#2 error for empathy \protect\subref{subfig:2} D\#2 error for distress \protect\subref{subfig:3} D\#3 error for empathy \protect\subref{subfig:4} D\#3 error for distress.}
\label{absolute error wrt true value}
\end{figure*}

\begin{figure*}[h]
\centering
\includegraphics[width=0.45\textwidth]{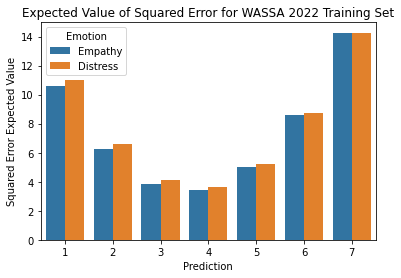}
\caption{Average squared error for predicted integer values on our training and validation data.}
\label{MSE expected value}
\end{figure*}

\begin{figure*}
\centering
\subfloat[]{\label{subfig:10}\includegraphics[height=6.5cm, width=10.5cm, keepaspectratio]{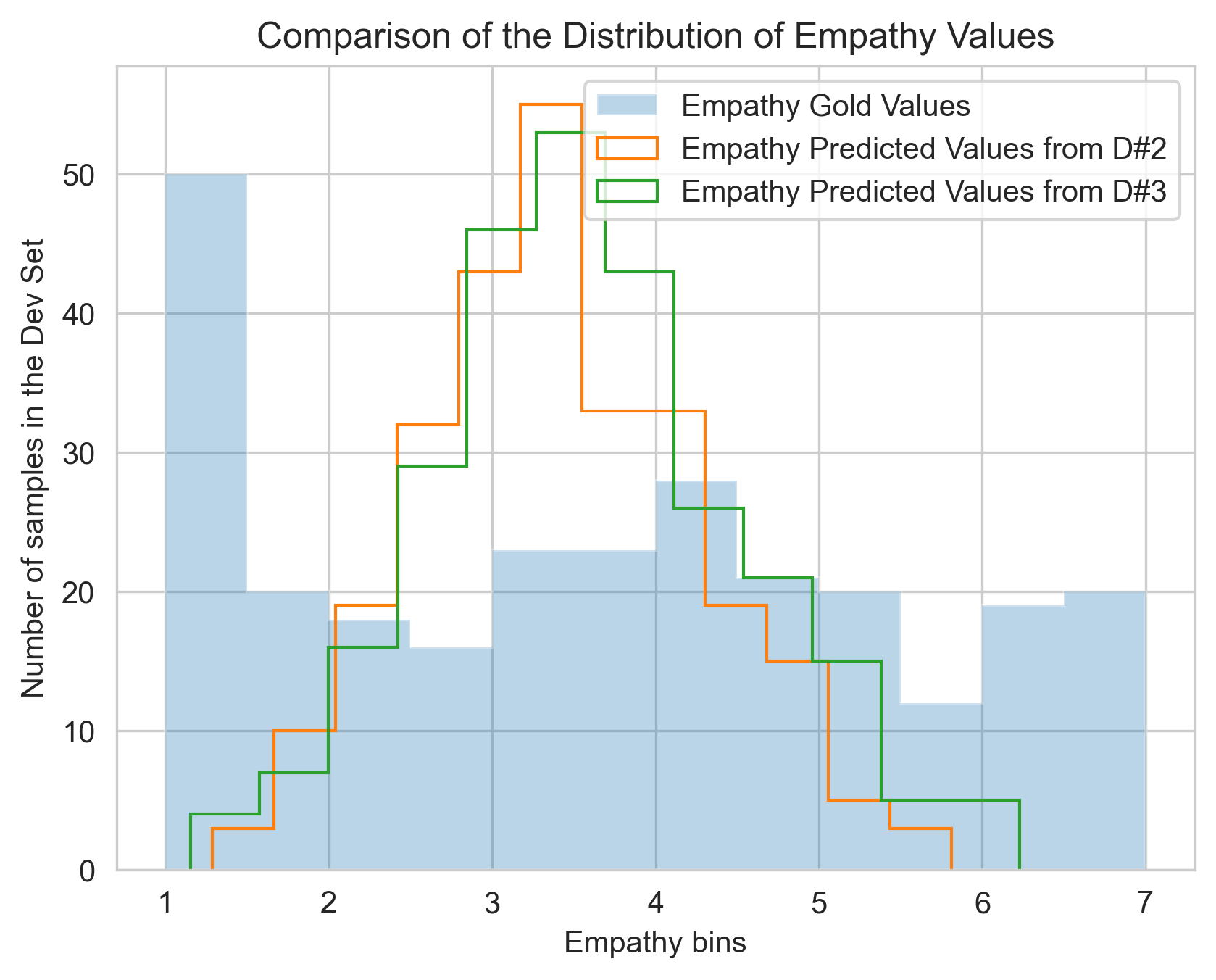}}
\subfloat[]{\label{subfig:11}\includegraphics[height=6.5cm, width=10.cm, keepaspectratio]{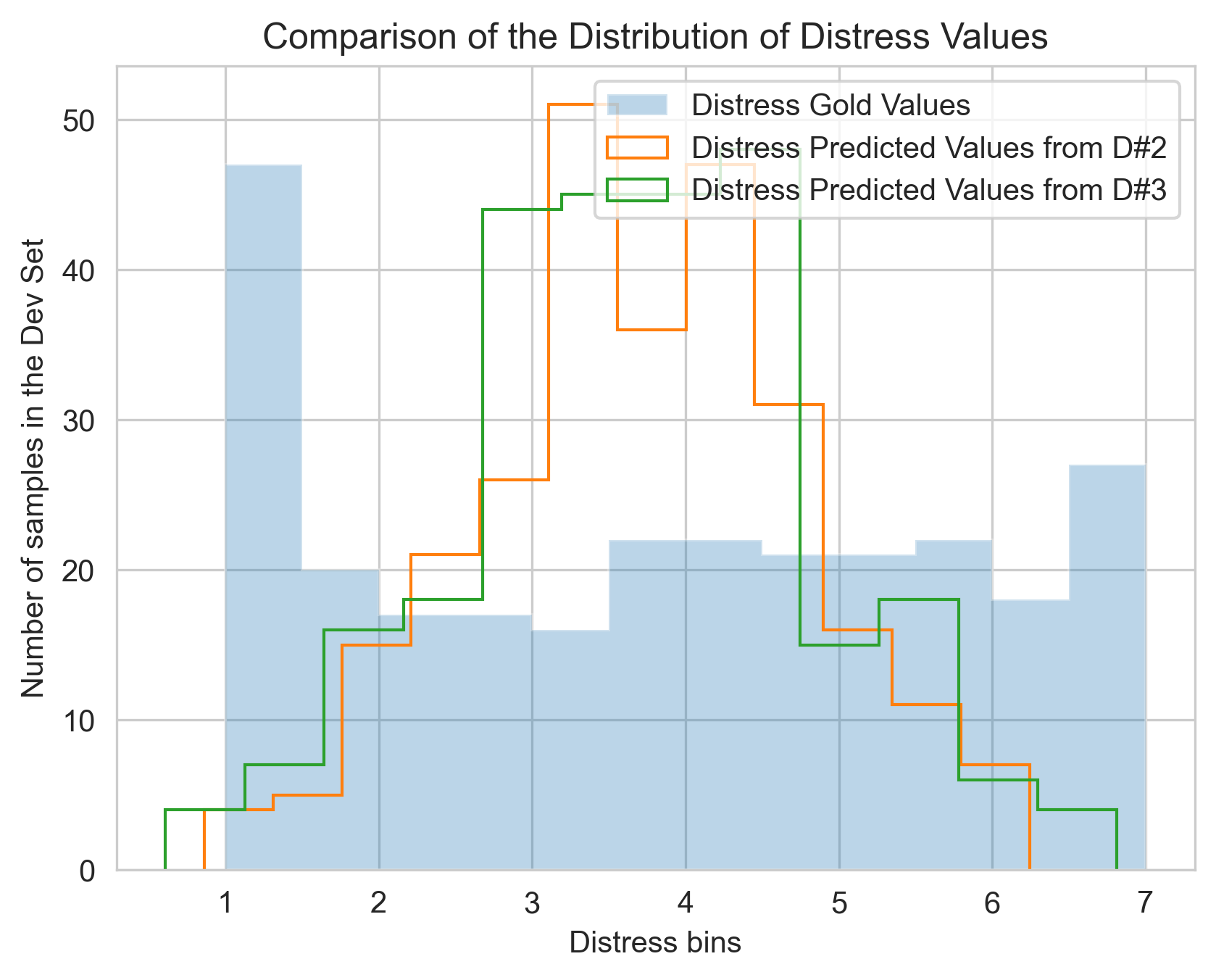}}
\caption{Comparison of gold and predicted values for the primary task:      \protect\subref{subfig:1} Empathy \protect\subref{subfig:2} Distress}
\label{dist_gold_predicted}
\end{figure*}

Error was also analyzed in relation to the distribution of values for empathy and distress. As seen in Figure~\ref{dist_gold_predicted}, the distribution of our models’ predictions varies from the distribution of the gold values for both empathy and distress. Table~\ref{center and spread measures} gives the mean and standard deviation of the gold values and our models’ predictions. While the center of our models’ predictions are fairly close to the means of the true distributions (the means of our predictions D\#2, D\#3 and D\#4 differ from the means of the gold values by less than 0.15), the standard deviations of our predictions are substantially smaller than the standard deviations of the gold values. Our models’ predictions favor values close to the center of the 1-7 range more frequently than our gold values do. This suggests that essays with emotion values at either end of the range may be more likely to have high error in our models’ predictions.

To test this hypothesis, absolute error was graphed with respect to true distress or empathy value in Figure~\ref{absolute error wrt true value}. The graphs in Figure~\ref{absolute error wrt true value} show a wide spread of absolute error values when emotion values are close to either end of the scale. Relative to the ends of the scale, the error values in the center are consistently low. As an additional test, the distance of each true value from the center of the scale was calculated, followed by the Pearson correlation between these distances and the absolute error. The Pearson correlations are shown in Table~\ref{centrality and error correlations}. For both emotions and all deliverables, distance from the center of the emotional scale has a moderate or strong positive correlation with the absolute error of our models’ predictions. Together, Figure~\ref{absolute error wrt true value} and Table~\ref{centrality and error correlations} confirm that our models perform better on essays whose true empathy or distress values are close to 4, and worse when true values are closer to 1 or 7.

Our analysis is that the calculation of loss during training causes our models’ differing performance on essays with true values at the center and ends of the scale. The expected value of squared error for integers 1 to 7 was calculated for the \citet{wassa-2022-codalab} training data and graphed in Figure~\ref{MSE expected value}. The average value of the squared error calculation is 6-10 squared units higher at the ends of the scale than in the center. This indicates higher average loss for predicting empathy and distress values at the ends of the scale may have guided our model toward predicting values close to the center of the scale more frequently.

It is worth noticing in Table~\ref{center and spread measures} that the center and spread of our D\#3 models’ predictions are closer to the true values’ center and spread than was the case for our D\#2 models. In particular, the standard deviation for both emotions for our D\#3 models’ predictions is higher than the standard deviations of the D\#2 models’ predictions by about 0.1. Compared to D\#2, our D\#3 models also have lower average absolute error on essays with true values within 1 unit of the ends of the scale (the average absolute error for empathy has improved by 0.049 and the average absolute error for distress has improved by 0.476, for essays where the relevant emotion has a true value less than 2 or greater than 6). This indicates that as we improved our approach during D\#3, the penalty for predicting values far from the center of the scale had less influence on our models’ predictions.

Our refinements to the primary task for D\#4 included an attempt to improve performance on data samples with true values close to 1 or 7 by assigning additional weight to those samples during the training of additional models within an ensemble, as discussed in Section \ref{section_primary_final_approach}. This attempt appears to have had some success. The Pearson correlation between absolute error and distances of true values from the midpoint of the scale (shown in Table~\ref{centrality and error correlations}) decreased from D\#3 by 0.219 for empathy and 0.116 for distress. While data samples far from the scale midpoint are still moderately positively correlated with absolute error for our D\#4 system, the strength of this correlation has decreased substantially from D\#3. The average absolute error for samples with true values within 1 unit of the ends of the scale also decreased (by 0.209 for empathy and 0.135 for distress). Finally, the standard deviation of our D\#4 system's predictions (shown in Table~\ref{center and spread measures}) increased by 0.212 for empathy and 0.195 for distress, moving closer to the standard deviation of the true labels. This means that our D\#4 system predicts values far from the midpoint of the scale more frequently than our previous models. Although data samples with true values near the ends of the empathy or distress scale are still a weakness for our final primary task system, performance on these samples continues the trend of improvement seen in our D\#3 model.

\subsection{Error analysis for the adaptation task} \label{section_adaptation_error_analysis}

\begin{figure*}
\centering
\subfloat[]{\label{subfig:16}\includegraphics[width=5.1cm, keepaspectratio]{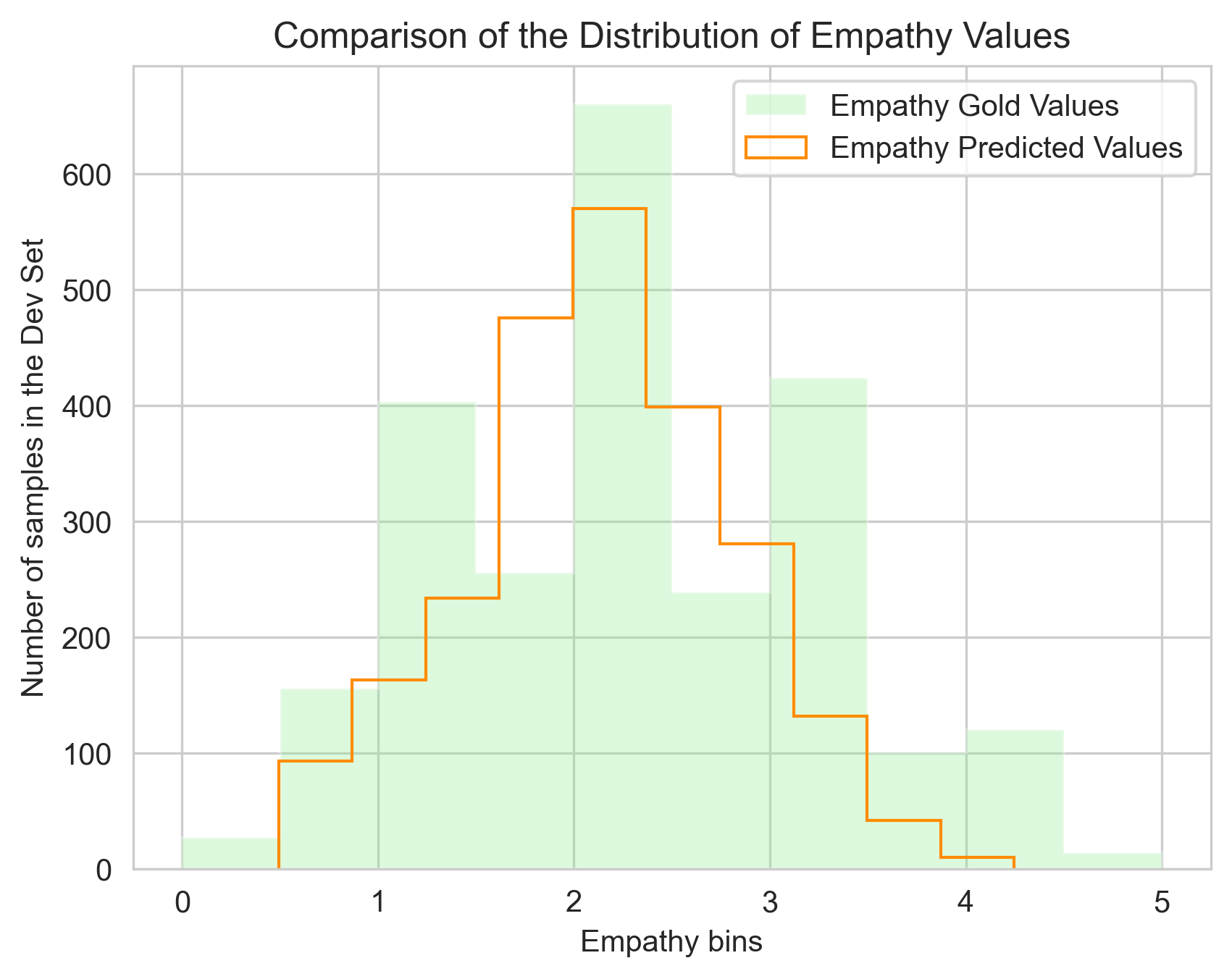}}
\subfloat[]{\label{subfig:17}\includegraphics[width=5.1cm, keepaspectratio]{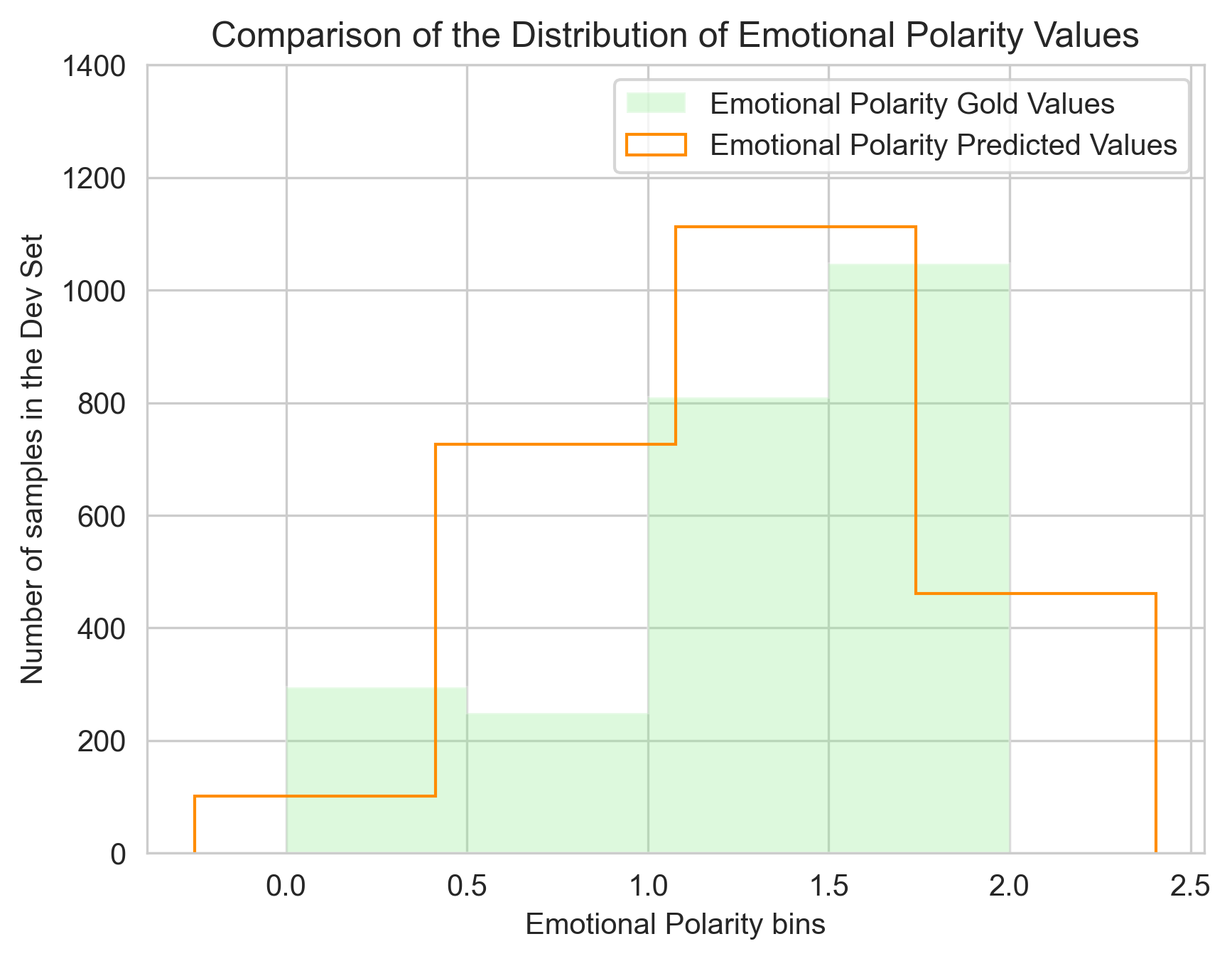}}
\subfloat[]{\label{subfig:18}\includegraphics[width=5.1cm, keepaspectratio]{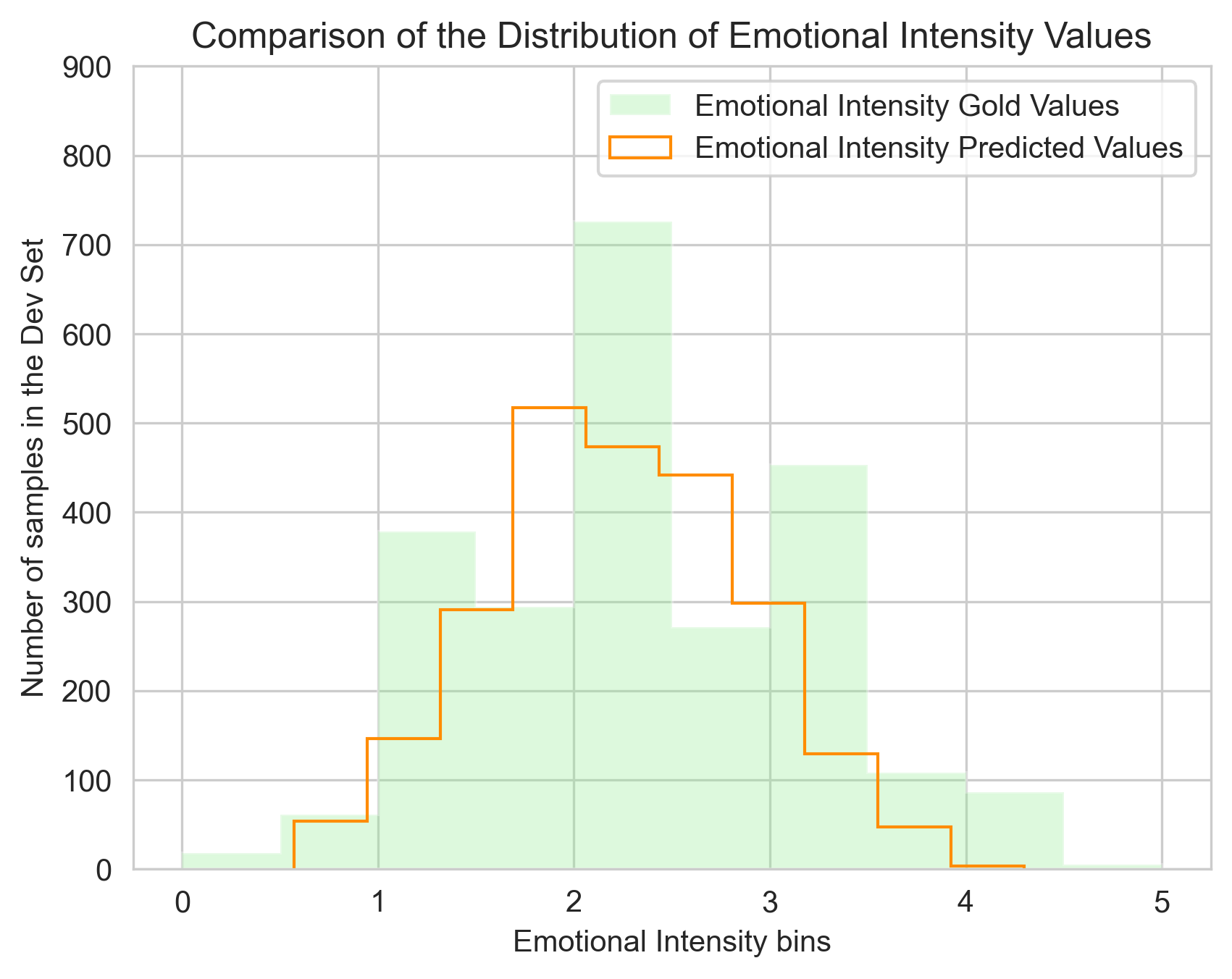}}
\caption{Comparison of gold and predicted values for the adaptation task:      \protect\subref{subfig:1} Empathy \protect\subref{subfig:2} Emotional Polarity \protect\subref{subfig:2} Emotional Intensity}
\label{dist_adaptation_predicted}
\end{figure*}

Figure~\ref{dist_adaptation_predicted} shows the comparison of distribution of target features against the gold values in the dev set. We can observe that each of the graphs show a high overlap between the predicted values and the gold values. This observation aligns with a relatively high Pearson coefficient for all these three predictions, with values of correlations over 0.7. 

We observe from the distribution of Empathy values in Figure~\ref{dist_adaptation_predicted} that the major prediction range is between 1.5 and 2. One example of high deviation is the conversation sample: ``And back to evidence - how can that attitude about life that is so clear and evident - be used in court these days? It seems so hard to trust anyone." The gold value of empathy for this sentence is 4, while the predicted value is 1.8. We believe that the reason for this large deviation is that the text is mixed with an emotion of anger. When observing the distribution of the predicted values of emotional polarity, we see that the prediction value spreads both towards negative polarity and positive polarity. Here is an example sentence: ``If I recall the response to the recent wildfire was kinda eh as well. Search and Rescue wise I mean. The firefighters worked very hard". While the gold standard emotional polarity of this sentence is 0, the predicted value is 1.8. We believe that the reason for this large deviation is that since the text is very high on positive emotion, the polarity would be high, which implies an accurate prediction. For emotional intensity, we observe that the predicted value shifts more towards the positive range. For example: ``mental illness is horrible and they deserve help". This sentence has the gold value of emotional intensity as 1.33 while the predicted value is 3.574. We believe the reason for this large deviation is that the text mixes both positive and negative emotion.

\section{Discussion}
The features for the D\#2 FFN models are high dimensional vectors of 384, 768, 1024, and 1536 dimensions for the different embedding models. During training we found that the models tended to overfit to the training data. Therefore, dropout layers and preserving the model weights using a validation set have been used to limit such effects. Although these changes resulted in a decrease in the rate at which the models overfitted, the issue remained prevalent at later epochs. These observations have been addressed in the second stage where an Advanced Dropout method has been used. The activation function has been updated and a learning rate scheduler has been implemented during the model training. These revisions are described in Section \ref{hyperparameter_tuning}. We can observe from Figure~\ref{plots_hyperparameter_before_after}, that these updates result in a stable training and validation loss. The plots of losses shows a smoother curve and the losses don't diverge even after training for over 100 epochs. We can also observe that the MSE loss values converge to stable levels and are still comparable to the best MSE losses obtained prior to hyperparameter tuning.

Furthermore, during the second stage, lexicon features have been added as described in Section \ref{section_lexicon}. As shown in the results in Section \ref{section_revised_results}, lexicon features added for the upgraded models result in an increase in performance. Their positive effect (examined through the correlation between lexicon features and empathy/distress scores) appears to come from two main sources: the NRC emotion lexicon and the MPQA subjectivity lexicon. Among these sources, higher counts in features related to negative feelings (e.g. sadness, fear, disgust, etc.), use of adjectives and weak subjectivity showed the biggest correlation to empathy/distress scores, as observed in Figure~\ref{correlation_lexicon_features}.

In the report for the first stage, we examined the distribution of the empathy and distress values in the designated development set, and we observed that the distribution is imbalanced, with a large spike for values between 1 and 2. However when we observed the distribution of empathy and distress values in the prediction, we find that the distributions appear to be Gaussian, peaking between 3 and 4. A comparative visual representation is shown in Figure~\ref{dist_gold_predicted}. During the second stage, we have revised the data sampling approach as detailed in Section \ref{section_sds}.

Figure~\ref{dist_gold_predicted} shows the cumulative effect of all the revisions performed during D\#3 on the predicted values. We can observe that although the predicted values retain a Gaussian distribution as we observed during D\#2, the predicted values have a wider distribution with lower peaks. These observations are in alignment with the results discussed in Section \ref{section_revised_results}.

The high-level approach developed for the primary task has been effectively adapted to the adaptation task. We can observe from Figure \ref{plots_adaptation} that the process for model training that was refined for the primary task, as described in Section \ref{hyperparameter_tuning}, has been successfully adapted to the adaptation task, and that the model training process is stable without overfitting. The combination of multiple approaches, namely, using Azure OpenAI embedding as described in Section \ref{section_embedding}, using lexicon features as described in Section \ref{section_lexicon}, and the neural network model architecture as described in Section \ref{section_adaptation_approach}, result in high predicted scores, as detailed in Section \ref{section_adaptation_results}. The CodaLab platform where the primary and adaptation tasks are hosted does not provide gold values of the test dataset for either of the tasks. Therefore a disaggregated analysis of the test data is not possible. However, a qualitative error analysis for the adaptation task has been performed on the dev dataset and is detailed in Section \ref{section_adaptation_error_analysis}. The finalized primary task for the the final stage scored 5th (as of 24th May 2023) position on the WASSA 2022 leaderboard.The model for the adaptation task scored 6th (as of 21st May 2023) position on the WASSA 2023 leaderboard.


\section{Ethical considerations}
\subsection{Dataset Usage}
The details of dataset and its license used in training of the model is updated in the dataset details in Section \ref{section_data_exploration}. 
\subsection{Essential elements for results reproducibility}
All the components such as dataset, code, and software requirement to reproduce the results are updated at the GitHub repository\footnote{\url{https://github.com/manisha-Singh-UW/LING573\_HUE-Human-Understanding-and-Empathy}}.

\section{Conclusion}

In the the first stage of this project, we had created an end-to-end functioning affect recognition system based on the WASSA 2022 Shared Task on Empathy Detection and Emotion Classification. The affect recognition system and associated approach used for this task are based on the teachings discussed in class and in readings. The designated development set from the shared task has been used to generate the output results and the scores of these results are based on the shared task's evaluation metric. The scores from the implementation using multiple embedding models have been presented in this report. 

The second stage builds upon the previous deliverable and presents three revisions, which have been described in the Section \ref{section_approach}. The Results Section \ref{section_results} describes the results of the revised system. The results also include the evaluation scores for the revised system, and compares them to the baseline system and also the results of the system presented during the first stage. 

As part of the final stage, we present the finalized end-to-end system for the primary task. We observe that the mean scores for the primary task on the EvalTest dataset increases by 33.59\% over the scores for the baseline. The approach developed for the primary task has been successfully adapted to the adaptation task, which is based on the WASSA 2023 Shared Task on Empathy Emotion and Personality Detection in Interactions. We observe an increase of 64.02\% in the mean score for the adaptation task on the EvalTest dataset when compared to the baseline.

\section{References}
\renewcommand{\section}[2]{}%
\bibliographystyle{acl_natbib}
\bibliography{acl2020}

\end{document}